\documentclass[11pt]{article}

\usepackage{cmap}
\usepackage{acl}

\usepackage{times}
\usepackage{latexsym}
\usepackage[T1]{fontenc}
\usepackage[utf8]{inputenc}
\usepackage{microtype}
\usepackage{inconsolata}

\usepackage{amsmath}
\usepackage{amssymb}
\usepackage{booktabs}
\usepackage{multirow}
\usepackage{graphicx}
\usepackage{xcolor}
\usepackage{float}

\newcommand{\coderepo}{https://github.com/XMUDeepLIT/WnW}

\newcommand{\blfootnote}[1]{%
  \begingroup\renewcommand\thefootnote{}\footnote{#1}%
  \addtocounter{footnote}{-1}\endgroup}

\title{WnW: Waxing-and-Waning KV Cache for Long-Form Speech LLMs}

\author{
  Yiming Yao\textsuperscript{1,2,*,\ddag} \quad
  Chenyang Lyu\textsuperscript{2,*} \quad
  Xuanfan Ni\textsuperscript{2} \quad
  Longyue Wang\textsuperscript{2} \\
  \bfseries Weihua Luo\textsuperscript{2} \quad
  Yazheng Yang\textsuperscript{1} \quad
  Jinsong Su\textsuperscript{1,\dag} \\[0.45em]
  \small\textsuperscript{1}School of Informatics, Xiamen University \quad
  \textsuperscript{2}Alibaba Group \\[0.2em]
  \small\textsuperscript{*}Equal contribution. \quad \textsuperscript{\dag}Corresponding author.
}

\begin{document}
\maketitle
\blfootnote{\textsuperscript{\ddag}Work done during an internship at Alibaba Group.}

\begin{abstract}
Long-form audio inputs make the KV cache the dominant memory cost of speech LLMs. Prefill-only KV compression methods permanently discard audio KV positions once evicted, with no pathway to recover them during decoding. We show this is fragile on long-form audio: prefill attention concentrates near the audio start (an attention-sink effect), while decode-time attention distributes broadly, and the two rankings overlap weakly. We propose \textbf{WnW (Waxing-and-Waning KV cache)}, which classifies KV-heads into \emph{anchor}, \emph{tidal}, and \emph{fixed} roles via offline calibration. Anchor heads keep all audio KV on GPU and yield a decode-time signal of which audio region each token is read from; tidal heads keep a CPU-resident complement that is recalled chunk-by-chunk based on aggregated anchor-head scores; fixed heads keep only an on-GPU subset, with the rest permanently discarded. On LibriSpeech-Long with two 3B backbones (Voxtral-mini-3b and Qwen2.5-Omni-3B), WnW preserves near-Full-Cache accuracy while keeping only $20\%$ of audio tokens on GPU, where prefill-only baselines fail to terminate. Results generalize across language, task, and domain shifts, and CPU--GPU recall adds little decode-time overhead in our measurements.
\end{abstract}

\section{Introduction}
\label{sec:intro}

Speech language models (LLMs) such as Voxtral~\citep{voxtral} now ingest audio inputs of arbitrary length and produce free-form text, enabling transcription of meetings, lectures, and long dialogues directly from raw waveforms, and increasingly speech translation as well~\citep{locatefocus,plast,csst}. Long-form audio shifts the inference bottleneck from the model's parameters to the key--value (KV) cache: at $12.5$--$25$ audio tokens per second, a ten-minute clip already occupies $7\,500$--$15\,000$ KV positions, and audio routinely accounts for $70$--$80\%$ of the total cache. Reducing the on-GPU audio KV footprint is therefore a prerequisite for deploying long-form speech LLMs under realistic memory constraints, and enables larger batch sizes on a fixed GPU budget.

KV cache compression methods for text LLMs, such as H2O~\citep{h2o}, SnapKV~\citep{snapkv}, and Ada-KV~\citep{adakv}, score every position once at prefill using attention from the prompt and retain the top-ranked positions throughout decoding. AudioKV~\citep{audiokv} adapts this paradigm to speech LLMs with FFT-smoothed scoring and an audio-tuned per-head budget. These methods share a common premise: prefill attention is a reliable proxy for the positions that decoding will query.

\begin{figure}[!t]
\centering
\includegraphics[width=\linewidth]{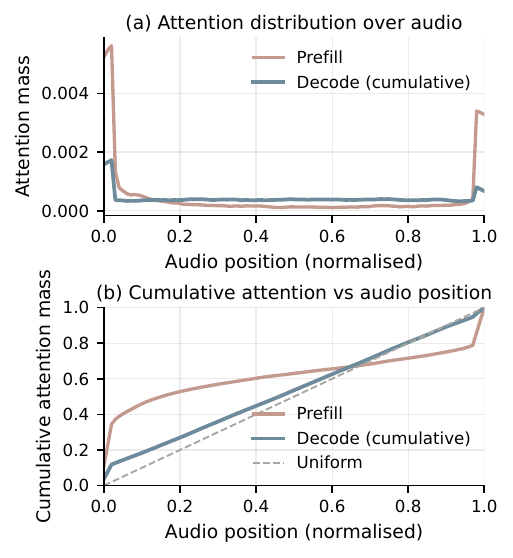}
\caption{Attention to audio positions, averaged across $50$ samples from the LibriSpeech-Long \texttt{dev-clean} subset (Voxtral). \emph{Prefill}: from the first answer token; \emph{Decode (cumulative)}: averaged over all answer tokens. (a) Per-position mass on a uniform grid; (b) cumulative mass. Prefill concentrates near the start (sink effect); decode-cumulative spreads across the full input.}
\label{fig:prefill_vs_decode}
\end{figure}

We find that this premise does not hold on long-form audio. On LibriSpeech-Long with Voxtral, prefill attention from the first answer token concentrates near the start of the audio (an attention-sink effect~\citep{streamingllm}), while the attention accumulated over the full decode distributes its mass broadly across the clip; the top positions selected by the two signals overlap weakly (Figure~\ref{fig:prefill_vs_decode}; \S\ref{sec:method:waxwane}). Any method that commits to a retention set early in inference and lacks a pathway to recover discarded positions is vulnerable to this gap; the failure is most visible for prefill-only baselines, which at tight budgets fail to terminate generation and produce word error rates well above 100\%.

We propose \textbf{WnW (Waxing-and-Waning KV cache)},\footnote{Code, experiment scripts, configurations, and dataset instructions will be released at \url{\coderepo}.} which reduces the on-GPU audio KV footprint while retaining recall of evicted positions from CPU. WnW classifies KV-heads into three functional classes via an offline calibration: a small set of \emph{anchor} heads is kept fully on GPU and serves as a decode-time observer of audio importance; \emph{tidal} heads keep a portion on GPU and the complement on CPU, available for recall; \emph{fixed} heads keep only an on-GPU subset. During decoding, anchor-head attention at each step is aggregated into chunk-level scores, the highest-scoring chunks are recalled from CPU into the tidal heads, and previously selected chunks that lose relevance are dropped from GPU. Deferring retention to decode time lets WnW correct prefill-time mistakes.

Our contributions are:
\begin{itemize}
\itemsep0em
\item We quantify the prefill/decode attention mismatch on long-form audio using positional mass concentration and top-$K$ Jaccard overlap, providing audio-domain evidence that any compression scheme without a recovery pathway is structurally limited.
\item We design WnW around two mechanisms that have no text-LLM counterpart. The head triage scores each head by how often its top-attended audio positions fall inside the word-aligned time span of the token being generated, multiplied by its gradient-based KV sensitivity, so the recallable tier is spent on the heads where discarding audio KV most damages quality. Because anchor heads are selected by exactly this criterion, their decode-time attention indicates which audio chunk the current token is read from, giving a recall signal aligned to audio time.
\item On LibriSpeech-Long with two 3B backbones (Voxtral-mini-3b and Qwen2.5-Omni-3B), WnW remains within $\sim1.6$ WER points of Full Cache even when only $20\%$ of audio tokens remain on GPU, a regime where prefill-only baselines fail to terminate. Results generalize across languages, tasks, and domains, scale favorably to a 24B backbone relative to KV-management baselines, and show limited CPU--GPU recall overhead.
\end{itemize}

\section{Method}
\label{sec:method}

\paragraph{Notation.}
Throughout, $L$ denotes the number of decoder layers, $H_q$ and $H_\text{kv}$ are the per-layer query and key/value heads (with GQA group size $g{=}H_q/H_\text{kv}$~\citep{gqa}), $r_\text{tok}$ is the audio frame rate in tokens per second, and $T_a$ is the number of audio KV positions in the input. For Voxtral-mini-3b-2507~\citep{voxtral} we have $L{=}30$, $H_q{=}32$, $H_\text{kv}{=}8$, $g{=}4$, and $r_\text{tok}{=}12.5$; for Qwen2.5-Omni-3B~\citep{qwen25_omni} we have $L{=}36$, $H_q{=}16$, $H_\text{kv}{=}2$, $g{=}8$, and $r_\text{tok}{=}25$.

\begin{figure*}[!tb]
\centering
\includegraphics[width=\linewidth,trim=0 25 0 29,clip]{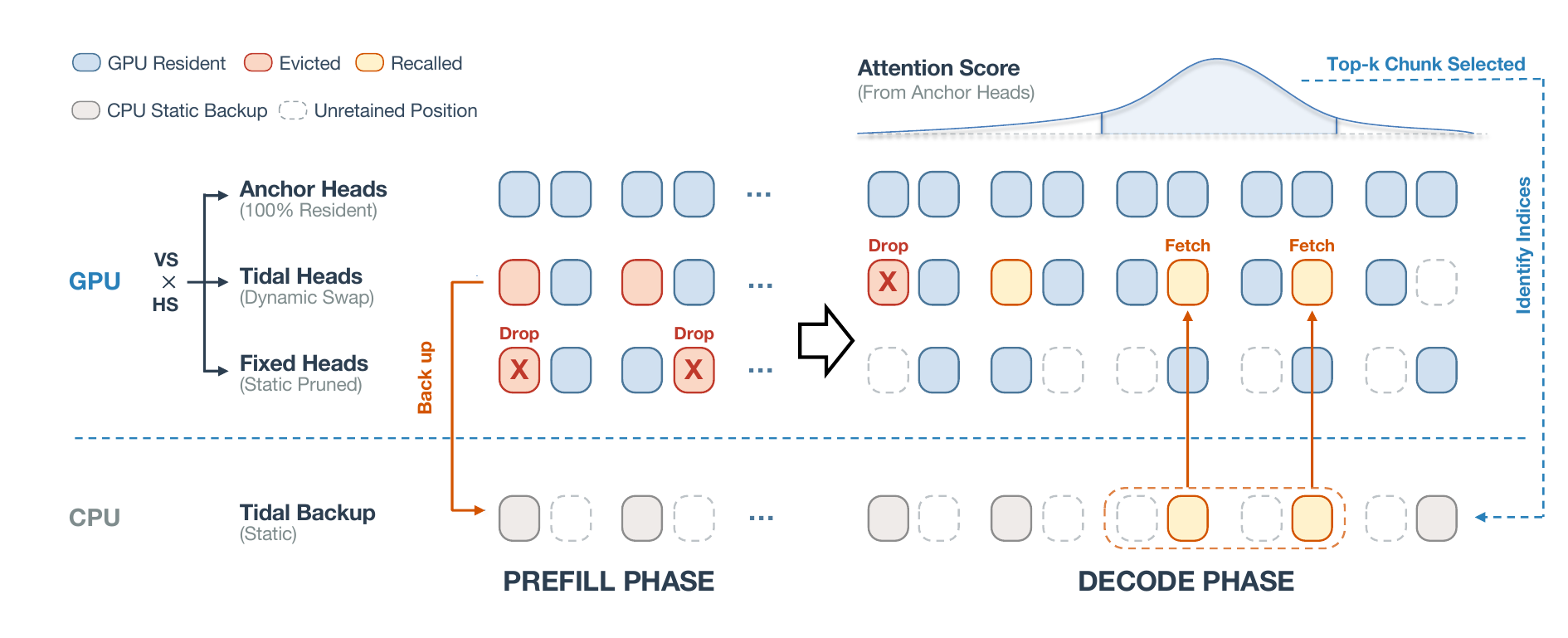}
\vspace{-0.8em}
\caption{Overview of WnW. \textbf{Left (Prefill)}: KV-heads are partitioned into three roles via offline VS$\times$HS calibration. Anchor heads retain all audio KV on GPU; tidal heads offload their unselected complement to CPU; fixed heads discard the unselected portion permanently. \textbf{Right (Decode)}: anchor-head attention identifies the top-$k$ audio chunks at each step; tidal heads recall the complement of those chunks from CPU on demand and release it again after $\delta$ consecutive unselected steps (the CPU copy persists). The dashed frame marks one selected chunk, i.e.\ two segments.}
\label{fig:method}
\vspace{-0.8em}
\end{figure*}

\paragraph{Method Overview.}
WnW compresses the audio KV cache through two coordinated mechanisms: an \emph{offline head triage} that partitions KV-heads into three functional roles, and an \emph{online recallable chunk swap} that tracks the generation frontier (Figure~\ref{fig:method}). The triage scores every KV-head on a held-out set along two complementary axes, voice score (VS) and head sensitivity (HS), then assigns each head to one of three roles: \textbf{anchor heads} keep all audio KV on GPU, and their attention indicates which audio region the current token is being read from; \textbf{tidal heads} keep a portion on GPU and offload the complement to CPU, recalling that complement on demand under anchor guidance; \textbf{fixed heads} keep a static per-segment prefill-time top-$k$ on GPU with no recall pathway.

\subsection{Prefill--Decode Attention Mismatch}
\label{sec:method:waxwane}

Methods that compress the audio KV cache at prefill, whether they were designed for text LLMs (SnapKV \citep{snapkv}, Ada-KV \citep{adakv}) or for speech LLMs (AudioKV \citep{audiokv}), score every audio position once, using attention from the prompt (or its last $w$ tokens) as a proxy for decode-time importance, and retain the top-ranked positions. This pipeline assumes prefill attention predicts decode-time attention. Text-domain work has tried to sharpen the prefill estimate via pseudo-query generation \citep{pseudoq} and learned future-attention adapters \citep{lookaheadkv}, acknowledging that the proxy is imperfect. We measure the gap directly on long-form audio.

Figure~\ref{fig:prefill_vs_decode} compares two attention distributions over audio positions, averaged across $50$ samples from the LibriSpeech-Long \texttt{dev-clean} subset (Voxtral). The \emph{prefill} distribution is the attention from the first answer token to each audio position, averaged over layers and heads; this is the signal token-level baselines use for compression decisions. The \emph{decode-cumulative} distribution additionally averages over all answer tokens produced during decoding. Two findings emerge:

\begin{itemize}
\itemsep0em
\item \textbf{Attention sink dominates prefill~\citep{streamingllm}.} The prefill distribution places $47.9\%$ of its mass on the first $10\%$ of audio positions and $69.3\%$ on the first half. The decode-cumulative distribution places only $9.8\%$ on the first $10\%$ and is close to uniform ($47.7\%$ / $52.3\%$ on the two halves).
\item \textbf{Top-$K$ rankings overlap weakly.} The Jaccard overlap between the top-$K$ audio positions ranked by prefill and by decode-cumulative attention is $0.187$ at $K{=}100$ and $0.240$ at $K{=}500$. This is well above the random baseline $\approx K/T_a$ ($0.013$ at $K{=}100$), but far from the agreement that would justify reusing prefill rankings during decoding.
\end{itemize}

Under these distributions, any method that fixes its retention set before decoding and lacks a recovery pathway will discard a substantial fraction of the positions decoding will query, irrespective of how that decision is scored. Extending the analysis to individual KV-heads, per-head Jaccard ($K{=}100$) ranges from $0.006$ to $0.641$ across the $240$ heads ($27\%$ below $0.1$): the mismatch is pervasive, but its quality impact is not, because heads also differ in how much the answer-token loss depends on their audio KV. The two factors compound: heads that both shift their attention as decoding progresses \emph{and} carry loss-critical audio KV need retention that evolves with the generated text; heads that fail either condition tolerate prefill's coarse approximation: without audio grounding, a head's top positions do not follow the current word through the clip, so the set prefill picks stays close to what decoding queries; without loss sensitivity, the answer-token loss barely depends on its audio KV, so quality is largely unaffected by which positions survive. WnW exploits this asymmetry: \S\ref{sec:method:triage} identifies the quality-critical heads via offline calibration, and \S\ref{sec:method:swap} invests recall overhead only in them.

Speech affords a structural advantage for decode-time correction: audio tokens are strictly time-ordered and word-aligned to the underlying transcript, so attention sweeps progressively through the audio as decoding advances. AudioKV \citep{audiokv} exploits this alignment at prefill to decide \emph{how much} to retain per head (a static budget), but leaves the \emph{which-positions} decision fixed. WnW takes the complementary step: it uses the temporal structure to decide \emph{when} to recall, letting anchor-head attention track the generation frontier and fetching the relevant audio chunk on demand. This is well-founded: self-attention captures sequence order through positional encodings and the training objective~\citep{yang2019wordorder}, both of which a transcription-trained speech LLM supplies, so anchor-head attention advances coherently through the time-ordered audio.

\subsection{Head Functional Triage via Offline Calibration}
\label{sec:method:triage}

In sequence-to-sequence learning, encoder layers differ sharply in importance, and generation quality hinges on a small subset that the decoder preferentially attends to~\citep{liu2021encoderfusion}. Audio KV-heads exhibit the same asymmetry, so we score each KV-head on a held-out set with two complementary signals: a \emph{voice score} (VS) measuring how strongly its attention is grounded in audio content, and a \emph{head sensitivity} (HS) measuring how much the answer-token loss depends on its audio KV. We combine them into modality-aware importance $\text{VS}_{l,k}\!\times\!\text{HS}_{l,k}$ that drives all triage decisions.

\paragraph{Voice Score (VS).}
We measure each head's \emph{audio-content grounding} as a hit ratio: the fraction of its top-attended audio positions that fall inside the word-aligned time span of the current answer token, with the alignment taken from WhisperX \citep{whisper,whisperx}. This construction follows AudioKV \citep{audiokv}, itself adapted from SparseMM \citep{sparsemm}. We average the ratio over answer tokens and calibration samples, then over the query heads that share each KV head, giving a per-KV-head voice score $\text{VS}_{l,k}$ (details in Appendix~\ref{app:calibration}).

\paragraph{Head Sensitivity (HS).}
Voice score captures attention patterns but not loss impact. We complement it with a gradient-based KV importance \citep{molchanov-taylor-pruning,michel-sixteen-heads}: $\text{HS}_{l,k}$ is the $\ell_2$-norm of the answer-token loss gradient with respect to each head's audio key vectors, averaged over positions, tokens, and calibration samples (Appendix~\ref{app:calibration}).

\paragraph{Three-Way Classification.}
We use the VS$\times$HS ranking over all $L \times H_\text{kv}$ KV-heads in two nested steps. First, the top $n_\text{voice}$ heads form the \textbf{voice heads}: those whose attention is grounded in audio content \emph{and} whose audio KV is loss-critical; their decode-time attention shifts as different words are generated, so they benefit from a recall pathway. The remaining heads are \textbf{fixed heads}: their decode-time attention pattern changes little after prefill, and we retain a static per-segment top-$k$ subset on GPU without recall. Second, within the voice heads, the top $n_\text{anchor}{=}5$ are designated \textbf{anchor heads} and keep their full audio KV on GPU: as the most audio-grounded and loss-critical heads, their decode-time attention is the most reliable per-step signal of which audio chunks matter, so we keep their KV uncompressed and reuse this signal to drive chunk swap (\S\ref{sec:method:swap}) for the tidal heads. The remaining voice heads are \textbf{tidal heads} and keep a portion on GPU with the complement on CPU.

Tidal and fixed heads share a proportional \emph{audio-KV} retention schedule:
\begin{equation}
\text{retention}_i = \min(\tilde{s}_i \cdot \lambda,\ 1.0),
\end{equation}
where $\text{retention}_i$ is the fraction of \emph{audio} positions kept on GPU for head $i$ (non-audio KV is always retained in full), and $\tilde{s}_i \in [0,1]$ is its normalized importance score. WnW exposes two hyperparameters: the voice-head count $n_\text{voice}$ controls how many heads have CPU-backed recallable storage, and the scale $\lambda$ controls the per-head audio retention magnitude. When $\lambda$ is large, all voice heads clip to retention $1$ and the recallable CPU partition is dormant; recall activates only when the GPU budget is tight enough to push $\lambda$ out of saturation.

\subsection{Prefill Compression and Decode-Time Chunk Swap}
\label{sec:method:swap}

\paragraph{Chunk Structure.}
We partition the $T_a$ audio positions into overlapping \emph{chunks} of $W_c{=}4$\,s with stride $W_s{=}2$\,s. A chunk holds $n_c = \lceil W_c \cdot r_\text{tok} \rceil$ audio tokens and spans two contiguous \emph{segments} of $n_s = \lceil W_s \cdot r_\text{tok} \rceil$ tokens, so adjacent chunks share a segment; load and eviction operate at the segment granularity. With $r_\text{tok}{=}12.5$ for Voxtral, this yields $n_c{=}50$ and $n_s{=}25$; with $r_\text{tok}{=}25$ for Qwen2.5-Omni, $n_c{=}100$ and $n_s{=}50$.

\paragraph{Prefill Compression.}
At prefill, each non-anchor KV-head $(l,k)$ ranks audio positions by aggregated text-to-audio attention $p^\text{prefill}_{l,k,\tau} = \sum_{t \in \text{text}} \text{attn}_{l,k,t,\tau}$. Selection is performed \emph{per segment}: within each $n_s$-token segment, the top $\lfloor \text{retention}_{l,k} \cdot n_s \rfloor$ positions are kept. This guarantees uniform temporal coverage, since no segment is wholly evicted, and provides a stable initial state for chunk swap; we ablate it against a global top-$k$ variant in \S\ref{sec:exp:ablation}. For tidal heads, the unselected positions are offloaded to CPU and remain recallable; for fixed heads, they are permanently discarded. Anchor heads keep all audio positions on GPU.

\paragraph{Decode-Step Score Aggregation.}
At every decode step $s$, after attention is computed for all $L$ layers, the anchor Q-heads' softmax weights at audio positions are aggregated across anchor (layer, q-head) pairs into a per-position score:
\begin{equation}
\begin{aligned}
A^{(s)}_\tau = \sum_{(l,h)\in\text{anchor}} & \mathrm{softmax}\!\left(\tfrac{q^{(s)}_{l,h}\, K_{l,h}^\top}{\sqrt{d}}\right)_\tau, \\
& \tau \in \text{audio}.
\end{aligned}
\end{equation}
The accumulator is reset to zero at the end of each step, so $A^{(s)}_\tau$ reflects only the current decode step, not a running sum across steps. This lets the importance signal track the generation frontier rather than being dominated by audio regions that mattered earlier.

\paragraph{Dynamic Chunk Selection.}
After each decode step, chunk $c$ receives score $\bar{A}^{(s)}_c = \text{mean}_{\tau \in c} A^{(s)}_\tau$; the top-$k$ ($k{=}3$) chunks are selected. For tidal heads, the positions selected at prefill remain GPU-resident throughout decoding, and what moves is their CPU-side complement: for a selected chunk, the complement of its segments is fetched from CPU onto GPU; a fetched complement left unselected for $\delta{=}3$ consecutive steps is dropped from GPU, while its CPU copy persists and can be re-fetched if the chunk regains importance.\footnote{The name \emph{WnW} (waxing and waning) refers to chunks moving across the GPU/CPU boundary as their per-step anchor-head attention rises and falls.} The eviction lag $\delta$ buffers short-term score fluctuations and prevents thrashing.

\section{Experiments}
\label{sec:exp}

\subsection{Models}
\label{sec:exp:models}

The main backbone is \textbf{Voxtral-mini-3b-2507} \citep{voxtral}, a 3B-parameter audio--language model; for cross-model evaluation we additionally use \textbf{Qwen2.5-Omni-3B} \citep{qwen25_omni} (per-model constants in \S\ref{sec:method}), whose feature extractor uses a single $300$-second window rather than the concatenated $30$-second windows of Voxtral, so it tests WnW across architectures rather than at unbounded input lengths; every clip we evaluate fits inside that window. Appendix~\ref{app:scaling} further reports a larger-scale check on \textbf{Voxtral-Small-24B}. All inference uses bf16 weights, greedy decoding, and the model's default chat template.

\subsection{Datasets}
\label{sec:exp:data}

\textbf{LibriSpeech-Long} \citep{librispeech_long} is a long-form benchmark constructed from LibriSpeech \citep{librispeech} by merging adjacent utterances into recordings of up to four minutes. We use \texttt{test-clean} and \texttt{test-other} (270 and 207 samples) as the main English ASR benchmark.

\textbf{LongSpeech} \citep{longspeech} is a multilingual long-form speech-LLM benchmark. We use its French ASR split (\texttt{asr-fr}) and English-to-French speech translation split (\texttt{en2fr}), sub-sampling 200 clips from each (fixed seed).

\textbf{PriMock57} \citep{primock57} is a set of $57$ simulated primary-care consultations, which we use as an out-of-domain ASR benchmark with its manual transcripts as references.

\textbf{Calibration set.} 50 LibriSpeech-Long \texttt{dev-clean} samples, disjoint from all test sets.

\subsection{Baselines}
\label{sec:exp:baselines}

We compare against five baselines, all using their authors' recommended hyperparameters unless noted (Appendix~\ref{app:hyperparams}):

\begin{itemize}
\itemsep0em
\item \textbf{Full Cache} --- no compression, the quality upper bound.
\item \textbf{Ada-KV} \citep{adakv} --- adaptive per-head budget allocation derived from a theoretical loss upper bound.
\item \textbf{AudioKV} \citep{audiokv} --- voice-score-guided per-head budget with FFT smoothing.
\item \textbf{ArkVale} \citep{arkvale} --- page-based recallable KV eviction with bounding-volume page summaries.
\item \textbf{AffPool} \citep{affpool} --- a token-merging efficiency baseline that pools adjacent audio tokens by affinity at prefill.
\end{itemize}

\paragraph{Audio-only compression constraint (modified baselines).}
We restrict the compression scope of all KV-management baselines to the audio-token region of the KV cache; text tokens (system prompt, user prompt, and tokens generated during decoding) are kept in full. The original implementations of Ada-KV, AudioKV, and ArkVale compress the entire sequence, which would mix audio compression with prompt and generated-token compression; restricting them to the audio region matches WnW's compression scope. AffPool needs no such restriction, since audio tokens are the only thing it ever touches.

The baselines span multiple axes: compression paradigm (static vs. recallable), domain (text vs. audio), and compression target (KV eviction vs. token merging). Each represents a broader family.\footnote{Ada-KV stands in for SnapKV \citep{snapkv}, PyramidKV \citep{pyramidkv}, HeadKV \citep{headkv}, RazorAttention \citep{razorattn}, and ChunkKV \citep{chunkkv}: however they allocate the budget, all fix their retention set at prefill with no recovery pathway, the property \S\ref{sec:method:waxwane} tests. ArkVale stands in for Quest \citep{quest}, ClusterKV \citep{clusterkv}, and InfiniGen \citep{infinigen}. KV quantization \citep{kivi} is orthogonal and compatible with WnW.} AffPool stands apart: it merges audio tokens before the KV cache exists, so it shrinks prefill computation as well as the cache, and its budget is the remaining audio-token ratio rather than a retention set over original positions.

\subsection{Audio KV Retention and WnW Configuration}
\label{sec:exp:budgets}

\paragraph{Audio KV retention.} Given the audio-only compression scope above, every method's compression budget is naturally expressed on the audio region. Let $N^\text{audio}_\text{full} = L \cdot H_\text{kv} \cdot N_\text{audio}$ be the size of the full per-(layer, KV-head) audio KV. We define
\[
r_\text{GPU} = \frac{N^\text{audio}_\text{GPU}}{N^\text{audio}_\text{full}}, \quad
r_\text{GPU+CPU} = \frac{N^\text{audio}_\text{GPU} + N^\text{audio}_\text{CPU}}{N^\text{audio}_\text{full}},
\]
where $N^\text{audio}_\text{GPU}$ and $N^\text{audio}_\text{CPU}$ are the audio KV token counts kept on device and (recallably) on host, respectively. $r_\text{GPU}$ is the on-device footprint that determines GPU memory; $r_\text{GPU+CPU}$ additionally counts any recallable CPU-resident audio KV. For methods without a CPU-resident complement (Ada-KV, AudioKV, Full Cache) the two coincide; for WnW (tidal heads) and ArkVale (evicted pages) they may differ.

\paragraph{Configuration.} We evaluate each method at four target retention levels $r_\text{GPU} \in \{0.8, 0.6, 0.4, 0.2\}$. Ada-KV, AudioKV, and ArkVale expose a single hyperparameter that maps directly to per-head GPU usage. For WnW, we select $\lambda$ on the calibration set so that the measured $r_\text{GPU}$ matches each target, using $n_\text{voice}\in\{30,30,50,90\}$ on Voxtral ($L{\times}H_\text{kv}{=}240$) and $\{9,9,15,27\}$ on Qwen2.5-Omni ($L{\times}H_\text{kv}{=}72$) for $r_\text{GPU}\in\{0.8,0.6,0.4,0.2\}$ respectively. $r_\text{GPU+CPU}$ equals $r_\text{GPU}$ down to $0.4$ and diverges only at $0.2$, where the recall pathway activates and it reaches $\approx 0.42$ (Table~\ref{tab:main}).

\subsection{Metrics}
\label{sec:exp:metrics}

\paragraph{Word Error Rate.} We report \emph{truncated WER} after light text normalization (lower-casing and removal of all punctuation except apostrophe and period): each hypothesis is first truncated to $1.2\times$ the ground-truth token length and then scored against the reference. Truncation prevents non-terminating hypotheses, which are common for prefill-only baselines at low budgets, from inflating insertion errors arbitrarily and isolates the effect of audio compression on the transcribed prefix. Substitutions/deletions/insertions are computed with the \texttt{jiwer} library.

\paragraph{Audio KV Retention.} For each method we report the \emph{measured} on-device retention $r_\text{GPU}$ (defined in \S\ref{sec:exp:budgets}), since the realized value may differ from the target by a few points. Tables~\ref{tab:generalization} and~\ref{tab:main} add WnW's combined retention $r_\text{GPU+CPU}$.

\subsection{Main Comparison}
\label{sec:exp:main}

\begin{figure}[!t]
\centering
\includegraphics[width=\linewidth]{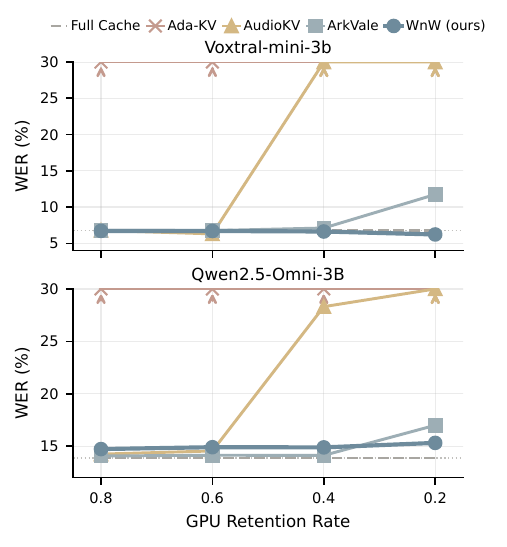}
\caption{Truncated WER vs.\ audio KV retention $r_\text{GPU}$ on LibriSpeech-Long \texttt{test-clean}. Top: Voxtral-mini-3b; bottom: Qwen2.5-Omni-3B. Upward arrows mark points exceeding $30\%$ WER (numbers in Table~\ref{tab:main}, Appendix). WnW stays close to Full Cache (dashed) at every $r_\text{GPU}$; prefill-only baselines degrade sharply below $r_\text{GPU}{=}0.6$.}
\label{fig:main}
\end{figure}

We compare WnW against the baselines on LibriSpeech-Long across the four target audio KV retention levels $r_\text{GPU}$, on both the Voxtral-mini-3b and Qwen2.5-Omni-3B backbones (Figure~\ref{fig:main}; full numbers in Table~\ref{tab:main}, Appendix).

Across both backbones, WnW stays close to the Full Cache at every retention level $r_\text{GPU}$, within $\sim$1.6 WER on Qwen2.5-Omni-3B. The baselines fall into a consistent ordering as $r_\text{GPU}$ decreases (in WER, lower is better): Ada-KV $\gg$ AudioKV $>$ ArkVale $>$ WnW. This ordering matches the prefill--decode mismatch analysis in \S\ref{sec:method:waxwane}: irreversible early compression (Ada-KV, AudioKV) inherits a structural error that grows with the eviction ratio, and both fail to terminate at tight $r_\text{GPU}$; adding a recall pathway (ArkVale) prevents non-termination but is insufficient alone; combining recall with audio-aware long-horizon scoring from anchor heads (WnW) closes the remaining gap. On Qwen2.5-Omni-3B, ArkVale slightly exceeds WnW at high retention: it keeps a complete CPU mirror of evicted pages ($r_\text{GPU+CPU}{=}100\%$), so $r_\text{GPU}$ controls only how much KV is GPU-resident, not how much audio stays recoverable, whereas WnW grants CPU recall only to tidal heads and permanently discards the unselected audio KV of fixed heads ($r_\text{GPU+CPU}{\approx}80\%$ at $r_\text{GPU}{=}0.8$). This is the intended storage--accuracy tradeoff of head triage: WnW stores less and is substantially better at low retention, where precise anchor-guided recall matters most. On Voxtral, WnW tracks the Full Cache within $\leq 0.6$ WER at every $r_\text{GPU}$, usually marginally below it; this gap is within the variation expected from greedy decoding and our truncation rule, not a real quality difference.\footnote{Our Qwen2.5-Omni-3B AudioKV result at $r_\text{GPU}{=}0.4$ is higher than reported by \citet{audiokv}. The official implementation is not public, so the gap may reflect implementation and evaluation differences, including head granularity under GQA, prompt template, \texttt{max\_new\_tokens}, and WER scoring. We apply one pipeline consistently to every method.}

\begin{table}[!t]
\centering
\footnotesize
\setlength{\tabcolsep}{3pt}
\begin{tabular}{@{}lccc@{}}
\toprule
Method & Category & Retention & clean / other \\
\midrule
Full Cache & --- & 100\% & \textbf{6.79} / \textbf{8.86} \\
\textbf{WnW} & KV recall & $\sim$20\% & \textbf{6.23} / \textbf{8.87} \\
AffPool & token merging & 20\% & 111.63 / 113.25 \\
AffPool & token merging & 40\% & 67.08 / 75.14 \\
AffPool & token merging & 60\% & 10.04 / 11.00 \\
AffPool & token merging & 80\% & 6.47 / 8.94 \\
\bottomrule
\end{tabular}
\caption{Comparison with a token-merging efficiency baseline on LibriSpeech-Long using Voxtral-mini-3b. Numbers are truncated WER ($\downarrow$, ratio $1.2$).}
\label{tab:affpool}
\end{table}

Table~\ref{tab:affpool} compares WnW with AffPool. AffPool recovers near-Full-Cache quality only at high retention, but collapses at $20\%$--$40\%$ retention. This suggests that prefill-time token merging shares the core limitation of the prefill-only baselines: it commits to its decision before decoding begins and leaves no way back. Merging adds an error-propagation effect on top, since merges made in early layers affect all subsequent layers. WnW avoids this failure mode by preserving token identities and deferring recall decisions to decode time. The KV-management ordering above also holds on the larger Voxtral-Small-24B backbone, where WnW remains the best of these methods at $\sim$20\% retention while the prefill-only baselines still collapse (Appendix~\ref{app:scaling}).

\subsection{Generalization and Domain Robustness}
\label{sec:exp:generalization}

To test whether the WnW design transfers beyond English ASR, we evaluate at the most aggressive setting $r_\text{GPU}{=}0.2$ on two LongSpeech splits, \texttt{asr-fr} (French ASR) and \texttt{en2fr} (English$\to$French speech translation), which together probe two orthogonal axes of generalization: cross-lingual and cross-task. We further test domain robustness on PriMock57 medical consultations (Table~\ref{tab:generalization}).

\begin{table}[!t]
\centering
\renewcommand{\arraystretch}{0.9}
\small
\setlength{\tabcolsep}{3pt}
\begin{tabular}{lcccc}
\toprule
& & \multicolumn{2}{c}{LongSpeech} & PriMock57 \\
\cmidrule(lr){3-4} \cmidrule(lr){5-5}
Method & $r_\text{GPU}$ & asr-fr$\downarrow$ & en2fr$\uparrow$ & WER$\downarrow$ \\
\midrule
Full Cache & 100\% & \textbf{20.42} & \textbf{38.48} & \textbf{23.47} \\
\midrule
Ada-KV     & $\sim$20\% & 112.91 & 2.99  & 106.43 \\
AudioKV    & $\sim$20\% & 105.00 & 5.22  & 103.83 \\
ArkVale    & $\sim$20\% & 28.63  & 35.40 & 32.45 \\
\textbf{WnW (ours)} & $\sim$20\% & \textbf{22.68} & \textbf{38.21} & \textbf{24.23} \\
\bottomrule
\end{tabular}
\caption{Generalization and domain robustness on Voxtral-mini-3b at $r_\text{GPU}{=}0.2$ (200 samples per LongSpeech split, all $57$ PriMock57 consultations, identical for all methods). asr-fr and PriMock57: truncated WER (ratio\,$1.2$); en2fr: sacreBLEU \citep{sacrebleu}, same truncation. WnW's measured retention is $r_\text{GPU}{=}22.25\%$ / $r_\text{GPU+CPU}{=}41.5\%$ on LongSpeech and $21.2\%$ / $40.46\%$ on PriMock57.}
\label{tab:generalization}
\end{table}

WnW transfers to a different language, a different task, and a different domain: relative to Full Cache it loses $2.3$ WER points on French ASR, $0.27$ BLEU on translation, and $0.76$ WER points on medical dialogue. Ada-KV and AudioKV again fail to terminate, producing hypotheses much longer than the reference and BLEU near zero. ArkVale avoids non-termination but trails WnW throughout, indicating that recall alone, without head triage, is insufficient. That the LibriSpeech-calibrated roles hold up under all three shifts suggests the partition is not tightly overfit to the calibration distribution. In a matched re-calibration study, replacing the English LibriSpeech calibration set with French LongSpeech changes English WER by at most $0.04$ points and French WER by $1.12$ points, with $76/85$ tidal heads preserved (Appendix~\ref{app:cross_calibration}).

\subsection{Ablation Studies}
\label{sec:exp:ablation}

We run all ablations on LibriSpeech-Long \texttt{test-clean} with truncated WER (ratio $1.2$). A1 and A2 are reported at both $r_\text{GPU}{=}0.1$ and $0.2$, since their effect depends on whether the recall pathway is active.

\textbf{(A1) Voice-head count $n_\text{voice}$.} We sweep $n_\text{voice}\in\{10,30,50,70,90\}$ at both retention levels, retuning $\lambda$ per cell so the on-GPU footprint matches the target. Since anchor heads stay at full retention regardless of $n_\text{voice}$, any quality change reflects the recall pathway: larger $n_\text{voice}$ converts fixed heads into CPU-backed tidal heads without enlarging the on-GPU budget.

\textbf{(A2) Prefill selection granularity.} We compare the default per-segment top-$k$ compression~(\S\ref{sec:method:swap}) against a global top-$k$ variant that spends the same total budget by ranking all audio positions of a head together with no per-segment quota. They share identical head triage, retention, and decode-time chunk swap, differing only in how prefill picks which positions to keep on GPU. We report at \texttt{nv30}/$r_\text{GPU}{=}0.1$ (tight) and \texttt{nv90}/$r_\text{GPU}{=}0.2$ (moderate).

\textbf{(A3) Head triage signal.} One signal selects both the top $n_\text{voice}$ voice heads and the top $5$ anchors, so we compare $\text{VS}$ alone, $\text{HS}$ alone, and $\text{VS}\times\text{HS}$ (WnW), holding everything else fixed at \texttt{nv30}/$r_\text{GPU}{=}0.1$.

\begin{table}[!t]
\centering
\renewcommand{\arraystretch}{0.85}
\small
\begin{tabular}{lcc}
\toprule
\multicolumn{3}{l}{\emph{(A1) Voice-head count (WER $\downarrow$)}} \\
$n_\text{voice}$ & $r_\text{GPU}{=}0.1$ & $r_\text{GPU}{=}0.2$ \\
\midrule
10 & 51.92 & 6.53 \\
30 &  9.78 & 6.46 \\
50 &  8.09 & 6.39 \\
70 & \textbf{6.31} & \textbf{6.20} \\
90 &  6.76 & 6.23 \\
\midrule
\multicolumn{3}{l}{\emph{(A2) Prefill selection (WER $\downarrow$)}} \\
 & \texttt{nv30}/$0.1$ & \texttt{nv90}/$0.2$ \\
\midrule
per-segment (ours) & \textbf{9.78} & \textbf{6.23} \\
global top-$k$     & 13.11 & 6.27 \\
\midrule
\multicolumn{3}{l}{\emph{(A3) Triage signal at \texttt{nv30}/$r_\text{GPU}{=}0.1$ (WER $\downarrow$)}} \\
\midrule
$\text{VS}\times\text{HS}$ (ours) & \multicolumn{2}{c}{\textbf{9.78}} \\
$\text{HS}$ only                  & \multicolumn{2}{c}{36.41} \\
$\text{VS}$ only                  & \multicolumn{2}{c}{124.66} \\
\bottomrule
\end{tabular}
\caption{Ablations on LibriSpeech-Long \texttt{test-clean} (truncated WER, ratio $1.2$).}
\label{tab:ablation}
\end{table}

A1 (Table~\ref{tab:ablation}) shows that larger tidal pools matter most under aggressive compression: at $r_\text{GPU}{=}0.1$, increasing $n_\text{voice}$ from $10$ to $70$ reduces WER from $51.92\%$ to $6.31\%$. At $r_\text{GPU}{=}0.2$ the same range moves WER by less than half a point, since the wider GPU budget already keeps most of what each head needs on device. A2 shows that per-segment prefill selection matters when $r_\text{GPU}$ is tight: at $r_\text{GPU}{=}0.1$, dropping the per-segment quota costs $3.33$ WER points, since global top-$k$ lets entire segments be evicted at prefill, leaving chunk swap no GPU-side starting point there; the gap closes to $0.04$ at $r_\text{GPU}{=}0.2$ when retention is high enough that uniform coverage is preserved either way. A3 shows that the two signals are not interchangeable under tight $r_\text{GPU}$: $\text{HS}$ alone trails $\text{VS}\!\times\!\text{HS}$ by $26.63$ WER points and $\text{VS}$ alone fails to terminate ($124.66\%$). Voice score by itself can promote audio-grounded heads whose KV is not loss-critical, while sensitivity alone can include heads that ignore the audio; only the product keeps both criteria. Appendix~\ref{app:sensitivity} further shows that WnW is insensitive to chunk size and anchor-head count around the default configuration.

\subsection{Efficiency and Runtime Analysis}
\label{sec:exp:efficiency}

\begin{figure}[!t]
\centering
\includegraphics[width=\linewidth]{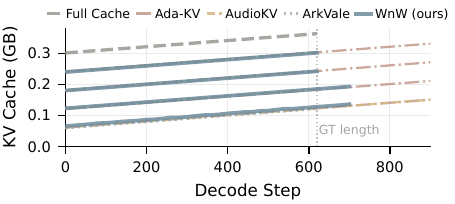}
\vspace{-1.6em}
\caption{GPU KV cache size over decode steps on a 4-minute LibriSpeech-Long sample (Voxtral-mini-3b). The four curves per method are $r_\text{GPU}\in\{0.8, 0.6, 0.4, 0.2\}$. The dotted vertical line marks ground-truth length; methods extending beyond it fail to terminate. WnW's slight oscillation reflects chunk swap.}
\label{fig:kv_growth}
\end{figure}

Figure~\ref{fig:kv_growth} shows the on-GPU KV cache size over decode steps under four compression ratios. All curves rise linearly because, per the audio-only constraint (\S\ref{sec:exp:baselines}), every method keeps generated tokens in full. At the same target $r_\text{GPU}$, all methods achieve nearly identical audio-KV footprints, so the WER differences in Figure~\ref{fig:main} reflect how each method uses its budget, not how much memory it consumes. Within this fixed GPU budget, WnW delivers near-Full-Cache accuracy down to $r_\text{GPU}{=}0.2$.

\begin{table}[!t]
\centering
\renewcommand{\arraystretch}{0.9}
\small
\begin{tabular}{lccc}
\toprule
Config. & $r_\text{GPU}$ & ms/token & MB/step \\
\midrule
35 tidal heads  & 21.7\% & 27.2 & 0.05 \\
75 tidal heads  & 21.4\% & 26.8 & 0.20 \\
115 tidal heads & 21.0\% & 26.7 & 0.40 \\
155 tidal heads & 20.5\% & 27.4 & 0.62 \\
195 tidal heads & 20.1\% & 27.5 & 0.82 \\
235 tidal heads & 19.7\% & 27.8 & 1.04 \\
\bottomrule
\end{tabular}
\caption{Runtime impact of CPU$\to$GPU recall on Voxtral-mini-3b, LibriSpeech-Long \texttt{test-clean} (64 samples) at $\sim$20\% GPU retention, on one A100 with batch size $1$. Decode time is median ms/token; MB/step is CPU$\to$GPU KV transfer averaged per decode step.}
\label{tab:runtime}
\end{table}

Table~\ref{tab:runtime} isolates recall cost by varying the number of tidal heads, the only heads that transfer from CPU. Transfer grows linearly with tidal-head count but stays small ($\leq1.04$ MB/step), and median decode time varies by less than $5\%$ across a $6.7\times$ range of tidal heads. CPU$\to$GPU recall is thus not the dominant bottleneck here.

\section{Related Work}
\label{sec:rw}

\paragraph{Static KV Cache Compression.}
Scissorhands \citep{scissorhands}, H2O \citep{h2o}, and SnapKV \citep{snapkv} retain a top-ranked subset of positions chosen from prefill or early-decode attention, and evicted positions are never re-admitted. Ada-KV \citep{adakv}, PyramidKV \citep{pyramidkv}, and ChunkKV \citep{chunkkv} refine per-layer, per-head, or per-chunk allocation under the same one-way eviction. All target text LLMs; \S\ref{sec:method:waxwane} shows their prefill-attention premise fails on long-form audio.

\paragraph{Head-Level KV Management.}
Some text-LLM methods allocate cache per head rather than per position. HeadKV \citep{headkv} scores heads for retrieval and reasoning ability to set per-head budgets, and RazorAttention \citep{razorattn} separates retrieval from non-retrieval heads, protecting the former in full while compressing the latter into a coarse compensation summary. WnW differs in both basis and consequence. The basis is audio grounding combined with gradient-based KV sensitivity rather than text-retrieval behavior, so head roles reflect how much discarding a head's \emph{audio} KV hurts quality. The consequence is that each head receives a storage tier rather than a cache budget: a tidal head keeps an exact copy of its unselected audio KV on CPU and can take it back unchanged mid-decode, whereas KV compressed into a summary is unrecoverable. Anchor heads further act as a decode-time importance observer, absent from static head-level schemes.

\paragraph{GPU--CPU KV Management.}
Building on paged GPU KV management \citep{vllm}, ArkVale \citep{arkvale} and ClusterKV \citep{clusterkv} evict KV pages or clusters and recall them on demand via bounding-volume or semantic digests; Quest \citep{quest} performs query-aware selective loading per decode step; InfiniGen \citep{infinigen} speculatively prefetches KV for upcoming steps. All target text LLMs and rely on per-step query similarity. WnW operates on time-aligned audio chunks, splits heads into functional roles with anchors as a dedicated importance observer, and aggregates attention across all anchor (layer, Q-head) pairs per decode step rather than scoring one query against page summaries.

\paragraph{Audio LLM KV Compression.}
AudioKV \citep{audiokv} adapts SnapKV-style static eviction to speech LLMs with FFT-smoothed scoring and an audio-tuned per-head budget; our voice score reuses its head-scoring hit ratio, itself adapted from SparseMM's visual score \citep{sparsemm}. WnW combines it with gradient-based head sensitivity \citep{molchanov-taylor-pruning,michel-sixteen-heads}, and drives recallable decode-time chunk swap rather than static prefill budgets. To our knowledge, this is the first recallable, decode-driven KV management for audio LLMs.

\section{Conclusion}
\label{sec:conclusion}

\looseness=-1
WnW addresses the prefill--decode attention mismatch on long-form audio by deferring retention decisions to decode time: an offline triage assigns anchor, tidal, and fixed roles, and tidal heads recall audio chunks from CPU under anchor guidance. It stays close to Full Cache on 3B backbones, generalizes across language, task, and domain, and leads KV-management baselines on 24B.

\section*{Limitations}
The chunk-swap CPU--GPU traffic is scheduled na\"ively; although our measurements show limited decode-time overhead, overlapping recall with attention via CUDA streams is a natural next optimization. The recall pathway is dormant in the high-$r_\text{GPU}$ regime (\S\ref{sec:method:triage}); decoupling $n_\text{voice}$ and $\lambda$ so that recall remains active when GPU is plentiful is a direction for further gains. Our evaluation targets offline decoding: extending WnW to streaming recognition is natural for speech-LLM architectures that append newly arrived audio to a shared, monotonically growing KV cache, as in recent LLM-based simultaneous speech translation systems that avoid re-encoding as input arrives~\citep{easist,fasst}; the same demand arises in the text-to-text case for LLM-based simultaneous machine translation~\citep{exposst}. The mechanism does not, however, transfer to cross-attention (e.g., Whisper), transducer, or state-space architectures, which keep no length-dependent audio KV cache to manage. Applying WnW to non-transcription audio tasks (QA, dialogue, summarization) is promising once the underlying speech LLMs reach reliable quality on those tasks.

\section*{Acknowledgments}
We thank the reviewers for their insightful comments.

\bibliography{custom}

\appendix

\section{Full Main Results}
\label{app:main_results}

Table~\ref{tab:main} (next page) reports the full LibriSpeech-Long results behind Figure~\ref{fig:main}, including \texttt{test-other} numbers and WnW's measured $r_\text{GPU}$ / $r_\text{GPU+CPU}$ at every retention level on both backbones.

\begin{table*}[!t]
\centering
\small
\begin{tabular}{lcccccccc}
\toprule
& \multicolumn{4}{c}{\texttt{test-clean} (WER$\downarrow$)} & \multicolumn{4}{c}{\texttt{test-other} (WER$\downarrow$)} \\
\cmidrule(lr){2-5} \cmidrule(lr){6-9}
Method ($r_\text{GPU}{=}$) & $0.8$ & $0.6$ & $0.4$ & $0.2$ & $0.8$ & $0.6$ & $0.4$ & $0.2$ \\
\midrule
\multicolumn{9}{c}{\emph{Backbone: Voxtral-mini-3b}} \\
\midrule
Full Cache  & \multicolumn{4}{c}{6.79} & \multicolumn{4}{c}{8.86} \\
\midrule
Ada-KV     & 137.46 & 175.85 & 196.01 & 199.24 & 144.01 & 181.15 & 192.60 & 198.55 \\
AudioKV    & 6.79   & 6.31   & 62.51  & 192.58 & 8.85   & 8.98   & 66.42  & 195.53 \\
ArkVale    & 6.81   & 6.77   & 7.12   & 11.74  & 8.83   & 8.85   & 9.79   & 14.75  \\
\textbf{WnW (ours)} & \textbf{6.72} & \textbf{6.71} & \textbf{6.64} & \textbf{6.23} & \textbf{8.84} & \textbf{8.84} & \textbf{8.81} & \textbf{8.87} \\
\midrule
\multicolumn{9}{l}{\emph{Measured audio KV retention}} \\
WnW $r_\text{GPU}$ (\%)            & 79.79 & 59.96 & 40.91 & 22.25 & 79.82 & 59.98 & 40.93 & 22.38 \\
WnW $r_\text{GPU+CPU}$ (\%)     & 79.79 & 59.96 & 40.91 & 41.52 & 79.82 & 59.98 & 40.93 & 41.64 \\
\midrule
\multicolumn{9}{c}{\emph{Backbone: Qwen2.5-Omni-3B}} \\
\midrule
Full Cache  & \multicolumn{4}{c}{13.87} & \multicolumn{4}{c}{16.80} \\
\midrule
Ada-KV     & 47.45  & 91.12  & 109.49 & 116.04 & 52.92  & 98.70  & 115.53 & 119.86 \\
AudioKV    & 14.24  & 14.54  & 28.31  & 131.30 & 17.62  & 17.58  & 34.16  & 128.27 \\
ArkVale    & 14.06  & 14.13  & 14.12  & 16.99  & 16.88  & 16.95  & 17.11  & 21.16  \\
\textbf{WnW (ours)} & \textbf{14.72} & \textbf{14.90} & \textbf{14.88} & \textbf{15.31} & \textbf{17.80} & \textbf{17.77} & \textbf{17.84} & \textbf{18.42} \\
\midrule
\multicolumn{9}{l}{\emph{Measured audio KV retention}} \\
WnW $r_\text{GPU}$ (\%)            & 80.26 & 60.12 & 40.23 & 21.78 & 80.31 & 60.17 & 40.28 & 22.00 \\
WnW $r_\text{GPU+CPU}$ (\%)     & 80.26 & 60.12 & 40.23 & 41.81 & 80.31 & 60.17 & 40.28 & 42.03 \\
\bottomrule
\end{tabular}
\caption{Full LibriSpeech-Long results across two backbones. Numbers are truncated WER ($\downarrow$, ratio $1.2$). The bottom rows of each backbone block report WnW's measured on-device retention $r_\text{GPU}$ and combined retention $r_\text{GPU+CPU}$ (on-device plus recallable CPU complement).}
\label{tab:main}
\end{table*}

\section{Scaling to a Larger Speech LLM}
\label{app:scaling}

We evaluate whether WnW's advantage over KV-management baselines persists beyond 3B-parameter backbones. Table~\ref{tab:scaling_24b} reports results on Voxtral-Small-24B at $\sim$20\% GPU retention. Head roles are recalibrated for the 24B model, while chunk size and anchor-head count are transferred from Voxtral-mini-3b.

\begin{table}[H]
\centering
\small
\begin{tabular}{lccc}
\toprule
Method & $r_\text{GPU}$ & test-clean & test-other \\
\midrule
Full Cache & 100\% & \textbf{5.70} & \textbf{8.37} \\
\textbf{WnW (ours)} & $\sim$20\% & \textbf{11.29} & \textbf{16.63} \\
ArkVale & $\sim$20\% & 15.60 & 22.20 \\
Ada-KV & $\sim$20\% & 110.51 & 113.05 \\
AudioKV & $\sim$20\% & 95.02 & 98.74 \\
\bottomrule
\end{tabular}
\caption{Scaling to Voxtral-Small-24B on LibriSpeech-Long. Numbers are truncated WER ($\downarrow$, ratio $1.2$).}
\label{tab:scaling_24b}
\end{table}

WnW remains the best KV-management method at $\sim$20\% GPU retention on the 24B backbone. The absolute gap to Full Cache is larger than on 3B models, reflecting the stronger base model and lower Full Cache WER, but prefill-only baselines collapse and ArkVale remains substantially worse than WnW.

\section{Hyperparameter Sensitivity}
\label{app:sensitivity}

Table~\ref{tab:sensitivity} reports sensitivity sweeps on Voxtral-mini-3b at $\sim$20\% GPU retention on LibriSpeech-Long \texttt{test-clean}. WnW is stable across a broad range of chunk sizes and anchor-head counts.

\begin{table}[H]
\centering
\small
\setlength{\tabcolsep}{4pt}
\begin{tabular}{@{}cc@{\hspace{1.2em}}cc@{}}
\toprule
\multicolumn{2}{c}{Chunk size} & \multicolumn{2}{c}{Anchor count} \\
\cmidrule(r){1-2}\cmidrule(l){3-4}
Value & WER & Value & WER \\
\midrule
0.96s & 6.45 & 1  & 6.22 \\
1.92s & 6.29 & 3  & 6.23 \\
4.00s (default) & 6.23 & 5 (default) & 6.23 \\
5.76s & 6.26 & 7  & 6.20 \\
7.68s & 6.23 & 10 & 6.21 \\
      &      & 15 & 6.23 \\
      &      & 20 & 6.22 \\
\bottomrule
\end{tabular}
\caption{Hyperparameter sensitivity at $\sim$20\% GPU retention on Voxtral-mini-3b, LibriSpeech-Long \texttt{test-clean}. Values are truncated WER ($\downarrow$, ratio $1.2$).}
\label{tab:sensitivity}
\end{table}

Across an $8\times$ range of chunk sizes and a $20\times$ range of anchor-head counts, WER fluctuates by no more than roughly $0.3$ points. This indicates that the default configuration is not the result of narrow-range tuning.

\section{Cross-Dataset Calibration Robustness}
\label{app:cross_calibration}

To directly test whether head roles overfit the calibration domain, we rerun the full calibration pipeline on 50 LongSpeech French ASR development samples and evaluate the resulting configuration against the original LibriSpeech-English calibration at the same operating point ($n_\text{voice}{=}90$, $n_\text{anchor}{=}5$, target $r_\text{GPU}{=}0.2$). Table~\ref{tab:cross_calibration} reports a matched re-run on English LibriSpeech-Long and French LongSpeech ASR.

\begin{table}[H]
\centering
\small
\setlength{\tabcolsep}{2.5pt}
\begin{tabular}{@{}lccccc@{}}
\toprule
& \multicolumn{3}{c}{WER $\downarrow$} & \multicolumn{2}{c}{Roles kept} \\
\cmidrule(lr){2-4} \cmidrule(l){5-6}
Calib. & clean & other & FR & anchor & tidal \\
\midrule
LS EN & \textbf{6.20} & \textbf{8.85} & \textbf{22.43} & --- & --- \\
LS FR & 6.24 & 8.89 & 23.55 & 2/5 & 76/85 \\
\bottomrule
\end{tabular}
\caption{Cross-dataset calibration robustness on Voxtral-mini-3b. Both configurations use WhisperX threshold $0.85$ and target $r_\text{GPU}{=}0.2$; numbers are truncated WER ($\downarrow$, ratio $1.2$). The \textsc{LS EN} row is an independent re-run of the default configuration, so it differs marginally from the same configuration in Table~\ref{tab:main}. \emph{Roles kept} reports how many of the anchor and tidal heads chosen by the English calibration are also chosen by the French calibration.}
\label{tab:cross_calibration}
\end{table}

Switching the calibration language and dataset changes English WER by at most $0.04$ points and French WER by $1.12$ points, while preserving most tidal heads. As a threshold sanity check, lowering the WhisperX confidence threshold from $0.85$ to $0.44$ increases the retained calibration tokens by $45\times$, yet the per-head voice-score correlation remains $0.974$ and WER changes by at most $0.31$ points relative to the French $0.85$ calibration. This suggests that the head roles reflect stable model behavior rather than a narrow calibration artifact.

Calibration is a per-model cost rather than a per-deployment one: head roles must be recomputed for a new backbone, as we do for Voxtral-Small-24B in Appendix~\ref{app:scaling}, but no shift in language, task, or domain in our results calls for recalibration. Recalibrating on French leaves the recall-eligible set largely intact, with 76 of 85 tidal heads shared, and does not improve French WER. Only 2 of 5 anchors coincide, but Appendix~\ref{app:sensitivity} shows that WER is nearly flat between one anchor and twenty, so the partition tolerates this reshuffling within the audio-grounded pool.

\section{Calibration Score Details}
\label{app:calibration}

\paragraph{Voice Score.}
For each answer token $t$, WhisperX forced alignment provides a word-level interval $[a_t, b_t]$, converted to audio-token indices $[\lfloor a_t \cdot r_\text{tok} \rfloor, \lfloor b_t \cdot r_\text{tok} \rfloor)$. The hit ratio for Q-head $(l, h)$ on token $t$ is
\begin{equation}
\text{hit}_{l,h,t} = \frac{1}{K} \sum_{k=1}^{K} \mathbf{1}\!\left[\text{topK}_{l,h,t}[k] \in [a_t^\text{tok}, b_t^\text{tok})\right],
\end{equation}
where $\text{topK}_{l,h,t}$ is taken over audio positions and $K$ is set to cover roughly one spoken word ($K{=}4$ for Voxtral, $K{=}8$ for Qwen2.5-Omni). Averaging over answer tokens (alignment confidence $\geq 0.85$) and $|D|{=}50$ calibration samples yields $\overline{\text{hit}}_{l,h}$, aggregated over GQA groups: $\text{VS}_{l,k} = \frac{1}{g}\sum_{j=1}^{g} \overline{\text{hit}}_{l,kg+j}$.

\paragraph{Head Sensitivity.}
\begin{equation}
S^\text{key}_{l,k,\tau} = \left\| \frac{\partial \mathcal{L}}{\partial K_{l,k,\tau}} \right\|,
\end{equation}
where $\mathcal{L}$ is the answer-token cross-entropy and $\tau$ indexes audio positions. The head sensitivity is
\begin{equation}
\text{HS}_{l,k} = \mathbb{E}_{D}\!\left[ \mathbb{E}_{t}\!\left[ \text{mean}_{\tau \in \text{audio}} S^\text{key}_{l,k,\tau} \right] \right].
\end{equation}

\section{Baseline Hyperparameters}
\label{app:hyperparams}

We use the authors' recommended hyperparameters for every baseline, summarized here.

\paragraph{Ada-KV \citep{adakv}.} \texttt{floor\_alpha}$=$0.8, kernel size 7, pooling \texttt{max}.

\paragraph{AudioKV \citep{audiokv}.} $\alpha{=}0.5$, $\rho{=}0.7$, observation window $32$, baseline ratio $0.5$. We use \texttt{min\_word\_score}$=$0.85 (rather than the original $0.95$) for both AudioKV and WnW so they share the same voice-score calibration; this is the only deviation from author defaults and applies symmetrically.

\paragraph{ArkVale \citep{arkvale}.} Page size $32$, top-$k$ recall budget proportional to the target audio KV retention $r_\text{GPU}$, CPU mirror of all evicted pages.

\paragraph{AffPool \citep{affpool}.} Deep pooling with window size $3$ is applied to the first $27$ layers of Voxtral-mini-3b, and input pooling with window size $1$ is applied to the last $3$ layers. The retention rate is controlled by calibrating the affinity threshold $\tau$ on a held-out subset, with measured retention within two percentage points of the target.

\section{AI Usage}
\label{app:ai_usage}

AI assistants were used only for linguistic refinement of the manuscript (clarity of arguments, logical coherence, and grammatical correctness) and for debugging our implementation code. The research idea, experimental design, and analysis of results were performed entirely by the authors, who assume full responsibility for the scientific content, conclusions, and integrity of this paper, and confirm compliance with academic ethics and the absence of plagiarism.

\end{document}